\documentclass[conference]{IEEEtran}
\ifCLASSOPTIONcompsoc
  \usepackage[nocompress]{cite}
\else
  \usepackage{cite}
\fi

\ifCLASSOPTIONcompsoc
  \usepackage[nocompress]{cite}
\else
  \usepackage{cite}
\fi
\ifCLASSINFOpdf
\else
\fi

\usepackage{color}
\definecolor{bluekeywords}{rgb}{0.13,0.13,1}
\definecolor{greencomments}{rgb}{0,0.5,0}
\definecolor{redstrings}{rgb}{0.9,0,0}

\usepackage{listings}
\usepackage[linesnumbered,ruled,vlined]{algorithm2e}
\usepackage{booktabs}

\usepackage{cite}
\usepackage{url}
\usepackage{parskip}
\usepackage{indentfirst} 
\usepackage{color}
\usepackage{graphicx}
\usepackage{caption}
\usepackage{subcaption}
\usepackage{diagbox}
\usepackage[linesnumbered,ruled,vlined]{algorithm2e}
\usepackage{mathtools}
\usepackage{amsmath}
\usepackage{verbatim}
\usepackage{pbox}
\usepackage{enumitem}

\newcommand{\unct}[1]{\ensuremath{\mathit{Uncertain}\langle \mathit{#1} \rangle}\xspace}

\newcommand{\uncT}{\unct{T}}

\begin{document}
\title{High Five: Improving Gesture Recognition by Embracing Uncertainty}

\author{
{\rm Diman Zad Tootaghaj$^\dag$, Adrian Sampson$^\ddag$, Todd Mytkowicz$^\ast$, Kathryn S McKinley$^{\ast\ast}$}\\
$^\dag$The Pennsylvania State University, $^\ddag$Cornell University, $^\ast$Microsoft Research, $^{\ast\ast}$Google Research \\
{\textit {\{dxz149\}@cse.psu.edu}, {\{asampson\}@cs.cornell.edu}, {\{toddm\}@microsoft.com}, {\{mckinley\}@cs.utexas.edu}}
}
\maketitle
\thispagestyle{plain}
\pagestyle{plain}
\begin{abstract}

  \noindent Sensors on mobile devices---accelerometers, gyroscopes,
  pressure meters, and GPS---invite new applications in gesture
  recognition, gaming, and fitness tracking. However,
  programming them 
  remains challenging because human gestures captured by sensors are noisy. This paper illustrates that noisy gestures
  degrade training and classification accuracy 
for gesture recognition 
in state-of-the-art
  deterministic Hidden Markov Models (HMM). We introduce a new
  \emph{statistical quantization} approach that mitigates these problems by
  (1) during training, producing
  gesture-specific codebooks, HMMs, and error models for gesture
  sequences; and (2) during classification, exploiting the error model
  to explore multiple feasible HMM state sequences. 
  We implement classification in \uncT, a probabilistic programming
  system that encapsulates HMMs and error models and then
  automates sampling and inference in the runtime. \uncT developers
  directly express a choice of application-specific
  trade-off between recall and precision at \emph{gesture recognition time},
  rather than at training time. We demonstrate 
  benefits in configurability, precision, recall, and recognition
  on two data sets with 25 gestures from 28
  people and 4200 total gestures. 
Incorporating
  gesture error more accurately in modeling improves the average
  recognition rate of 20 gestures from 34\% in prior work to
  62\%. Incorporating the error model during classification \emph{further}
  improves the average gesture recognition rate to 71\%. As far as we are
  aware, no prior work shows how to generate an HMM error model during
  training and use it to improve classification rates.



\end{abstract}

\section{Introduction} \label{introduction}
\noindent Modern mobile devices host a diverse and expanding array of sensors:
accelerometers, gyroscopes, pressure meters, thermometers, ambient
light sensors, and more. These sensors invite new experiences in fitness, health, translating sign language, games, and accessibility for people with disabilities~\cite{liang1998real, starner1998real, hinckley2003synchronous}. Despite all these new input methods, user input on smartphones is still mostly limited to touching the screen and keypad, a 2-D detection problem. This paper
identifies and addresses algorithmic and practical impediments to deploying 3-D gesture recognition on smartphones. We extend the commonly used Hidden Markov Model (HMM) approach~\cite{ghahramani, lee, schlomer, zhang2009hand}.  Determining a 3-D path through space is harder than 2-D gesture recognition~\cite{wobbrock2007gestures} because human gestures as captured by sensors are uncertain and noisy---much noisier than the sensors themselves. Humans hold the device at different angles, get tired, and change their gestures' pattern.   
Prior state-of-the-art gesture-recognition algorithms using
HMMs~\cite{schlomer, mantyjarvi2004enabling, hofmann1998velocity} are
limited because (1) they assume all gesture error is uniform and project all observations to one spherical codebook for HMM training; and (2) classification generates one observation sequence and produces only one deterministic gesture, rather than reasoning explicitly about the uncertainty introduced by gesture error.

We measure 
gesture noise in accelerometer
data and find it is a gesture-specific Gaussian mixture model: the error
distributions along the x, y, and z axes all vary. In contrast, when
the phone is still, accelerometer error is extremely small, Gaussian,
and uniform in all dimensions.  
Gesture-specific error matches our intuition about humans. Making an
``M'' is harder and more subject to error than making an ``O'' because
users make three changes in direction versus a smooth
movement.  Even when making the same gesture, humans hold and move
devices with different orientations and rotations. Since
gesture observation is a sequence of error readings, differences in
gesture sizes and speed can compound gesture error.

State-of-the-art HMM systems~\cite{schlomer, mantyjarvi2004enabling, liu2009uwave} assume errors are small, uniform, and not gesture specific. They compute one radius for all gestures and all x, y, and z accelerometer data. They map all gestures to a single spherical set of codewords centered at $(0,0,0)$ that they use to train the HMM.
Classification compounds this problem because HMMs use \emph{deterministic
quantization}. Even though several nearby states may be very
likely, traditional HMM classifiers only explore one.

To solve these problems, we present a holistic \emph{statistical quantization} approach that (a) computes and reasons about noise
gesture training data; (b) produces per-gesture HMMs and
their error models; (c) modifies classification to
use the error model to choose the most likely gesture; and (d) uses
the \uncT probabilistic programming
system~\cite{bornholt2014uncertain, uncertain:uncertainT} to simplify
the implementation and expose the classifier's trade-off between
precision and recall.


During training, we measure error in accelerometer data sequences
across gestures and use the mean and variance to improve HMM modeling and classification. In training, we fit per-gesture data to codewords on an ellipse and generate gesture specific HMM codebooks. We show that ellipse-based codebooks improve accuracy over prior sphere-based approaches~\cite{schlomer, mantyjarvi2004enabling}.
We target personal mobile devices where users both specify and train
gestures. With per-gesture HMM models, 
users train one gesture at a time. Instead of performing classification by deterministically mapping the 3-D acceleration data to the closest codeword, we sample from the error model produced during training to explore a range of
potential gestures.  

We implement classification as a library in the \uncT
programming language. The library provides trained HMM models and their error models. A gesture is an Uncertain
type. 
Values of Uncertain types represent probability distributions by
returning samples of the base type from the error distribution. The
runtime lazily performs statistical tests to evaluate computations on
these values. 
When the application queries an Uncertain value, such as with an
\texttt{if} statement on the gesture, the runtime performs the specified statistical hypothesis test by sampling values from the HMM computation.
 
We evaluate statistical quantization on two data sets: (1) five
gestures trained by 20 people (10 women and 10 men) on a Windows Phone
that we collect, and (2) 20 gestures trained by 8 people from Costante
et~al.~\cite{costante2014personalizing}. Compared to traditional
deterministic spherical quantizers~\cite{schlomer}, statistical
quantization substantially improves recall, precision, and recognition
rate on both data sets. Improvements result from better modeling and using error
in classification. \emph{Deterministic elliptical quantization} improves average gesture recognition rates on 20 gestures to 62\%, compared to the 34\% for traditional \emph{deterministic spherical quantization}. \textit{Statistical elliptical quantization} further improves gesture recognition rates to 71\%.  

We illustrate the power of our framework to trade off
precision and recall 
because it exploits the error model during classification. Prior work
chooses one tradeoff during training. Different configurations significantly improve both precision and recall. This capability makes statistical quantization suitable both for applications where false positives are undesirable or even dangerous, and for other applications that prioritize making a decision over perfect recognition. 

\emph{Our most significant contribution is showing how to derive and use gesture error models to improve HMM classification accuracy and configurability.} Our
\uncT approach is a case study in how a programming
language abstraction for error inspires improvements machine-learning
systems.  HMM inference algorithms, such as Baum--Welch, exemplify
software that ignores valuable statistical information because it can
be difficult to track and use.  They infer a sequence of hidden states
assuming \emph{perfect} sensor observations of gestures. 
Our
approach invites further work on enhancing inference in other machine learning
domains, such as speech recognition and computational biology, that
operate on noisy data.  We
plan to make the source code available upon publication.  The \uncT
compiler and runtime are already open
source~\cite{uncertain:uncertainT}.


\section{Overview of Existing Approaches} \label{Background}
\begin{figure}[!tb]
\centering
\includegraphics[height=1.5in, width=3.5in]{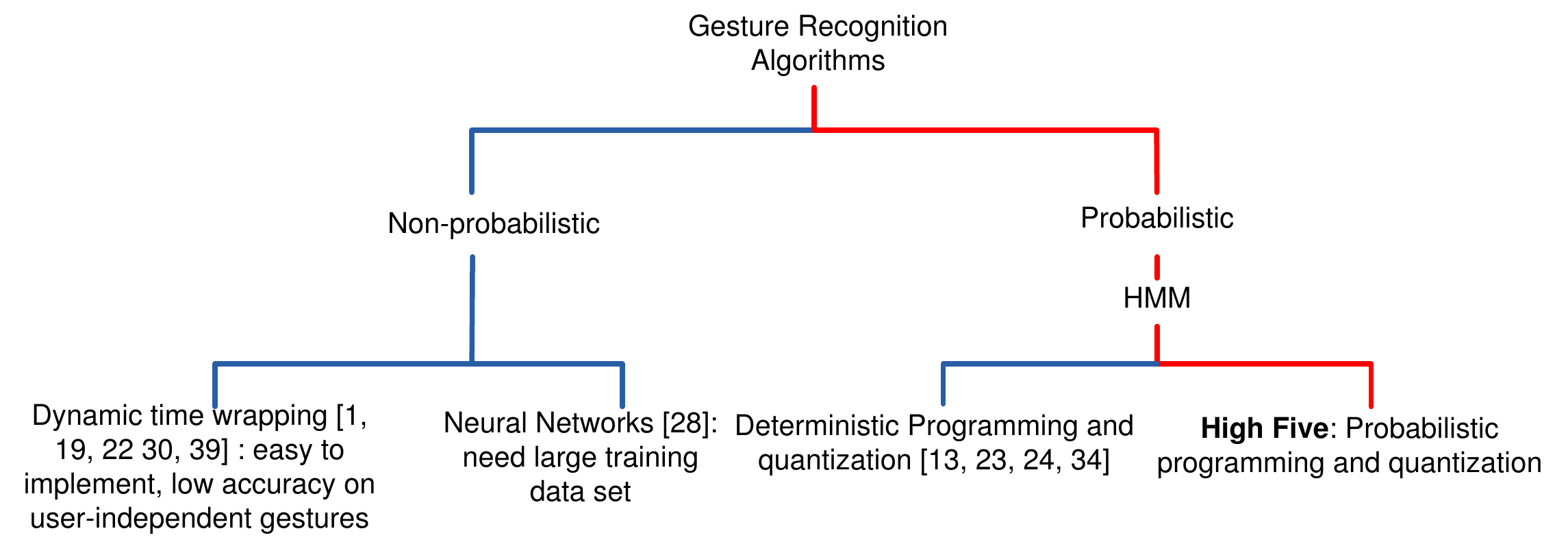} 
\caption{\label{fig:Related} Design space for gesture recognition algorithms.}
\end{figure}
Recognizing human gestures is key to more natural human-computer interaction~\cite{mitra, schlomer, akl2010accelerometer, niezen2008gesture}. Sensor choices include data gloves \cite{liang1998real}, cameras~\cite{zhang2012microsoft}, touch detection for 2-D painting gestures \cite{wobbrock2007gestures}, and our focus, 3-D motion tracking accelerometer and gyroscope sensors \cite{kristensson2012continuous}. 
Figure~\ref{fig:Related} presents the design space for common gesture recognition approaches. Non-Probabilistic approaches include dynamic time warping \cite{akl2010accelerometer, niezen2008gesture, liu2009uwave, wilson2004gesture, mace2013accelerometer} and neural networks \cite{murakami1991gesture}. A common probabilistic approach is Hidden Markov Models (HMMs) \cite{schlomer, mantyjarvi2004enabling, mantyla2001discrete, hofmann1998velocity} that use non-linear algorithms to find similar time-varying sequences.\\ 
\indent \textbf{Hidden Markov Models}: HMMs for gesture recognition give the best recognition rates for both user-dependent and user-independent gestures
\cite{ali2014comparative}. 
Our HMM implementation for gesture recognition differs from the prior literature as follows. First, instead of using a deterministic codebook for discretization, we use the statistical information about each gesture during HMM training and generate a different codebook for each gesture. Second, we exploit a probabilistic programming framework in our implementation and use uncertain data types to make more accurate estimations of the probability of each gesture. Third, unlike prior work that deterministically maps raw data to one static codebook, we use a stochastic mapping of raw scattered data based on the gesture's error model and the distance from the data to each gesture's trained codebook.\\
\indent \textbf{Kmeans quantization}: Since continuous HMMs for gesture recognition is impractical due to high complexity of tracking huge observation states, a variety of quantization techniques transform sensor data into discrete values. The most-common is \textit{kmeans} clustering \cite{schlomer, hartigan1979algorithm}. Although \textit{kmeans} works well for large isotropic data sets, it is very sensitive to outliers and therefore noisy sensor data degrades its effectiveness. For example, a single noisy outlier results in a singleton cluster.  Furthermore, because humans must train gesture recognizers, the training gesture data sets 
are necessarily small, whereas \textit{kmeans} is best suited for large data sets. \\
\indent \textbf{Dynamic time warping}: Dynamic time warping applies dynamic programming to match time-varying sequences where gesture samples are represented as feature vectors \cite{akl2010accelerometer, niezen2008gesture, ali2014comparative}.  The algorithm constructs a distance matrix between each gesture template $T=\{t_1, t_2, ...\}$ and a gesture sample $S=\{s_1, s_2, ...\}$. The algorithm next calculates a matching cost $\mathrm{DTW}(T,S)$ between each gesture template and the sample gesture. The sample gesture is classified as the gesture template with the minimum matching cost. This approach is easy to implement, but its accuracy for user-independent gestures is low \cite{liu2009uwave, paudyal2016sceptre}. Furthermore, it is deterministic and does not capture the stochastic and noisy behavior of accelerometer and gyroscope's data.\\
\indent \textbf{Neural networks}: Neural networks classify large data sets effectively \cite{murakami1991gesture}. 
During training, the algorithm adjusts the weight values and biases to improve classification. While neural networks work well for large data sets, their applicability is limited for small amounts of training data \cite{lee}. Asking end users to perform hundreds of gestures to train a neural network model is impractical.
\begin{figure}[!tb]
\centering
\includegraphics[height=2in, width=3in]{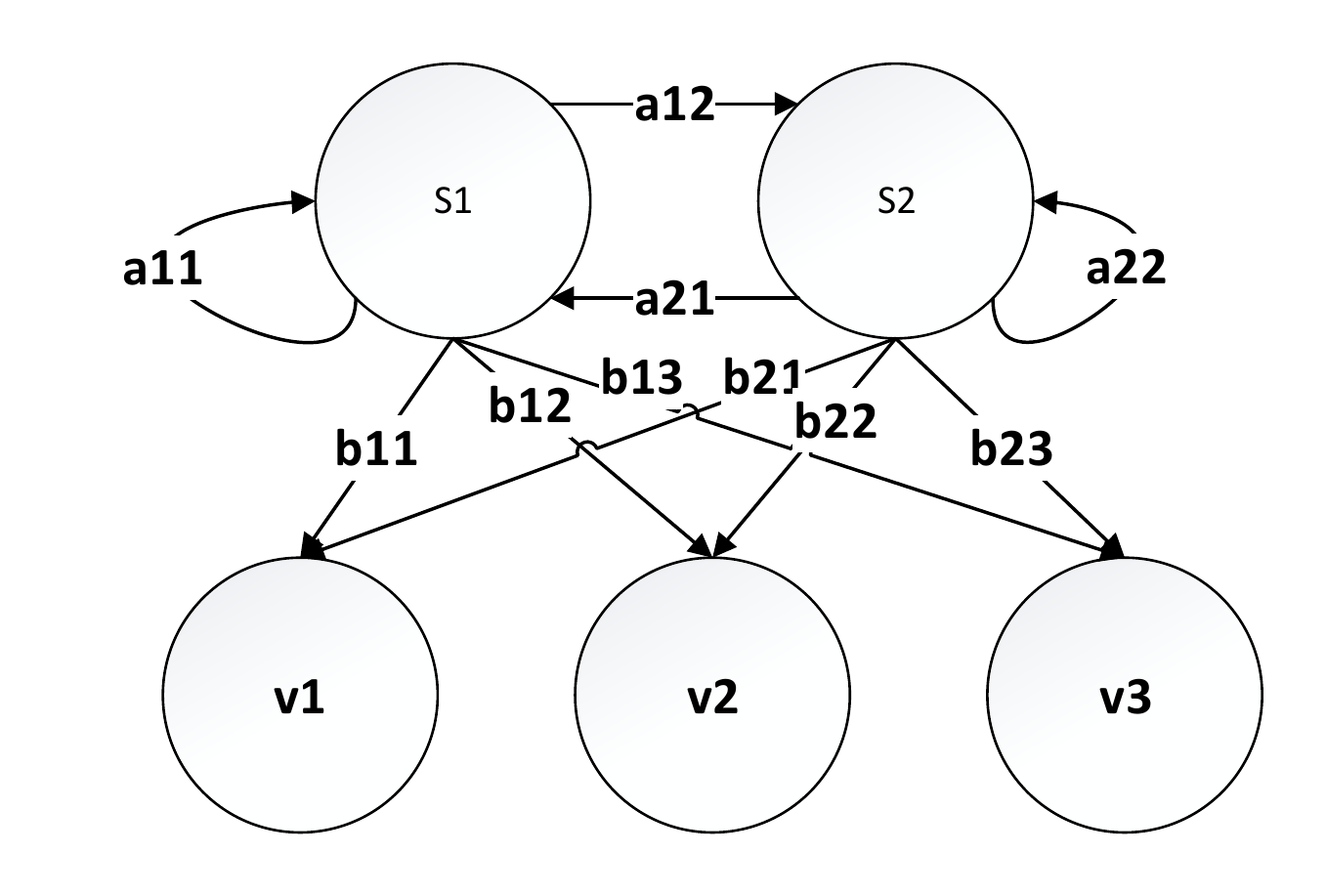} 
\caption{\label{fig:HMM} HMM example with state and output probabilities.}
\end{figure}
\section{Hidden Markov Model Background}
HMMs are used extensively in pattern recognition algorithms for speech, hand gestures, and computational biology. HMMs are Bayesian Networks with the following two properties: 1) the state of the system at time t ($S_t$) produces the observed process ($Y_t$), which is a random or deterministic function of the states, is hidden from the observation process, and 2) the current state of the system $S_t$ given the previous state $S_{t-1}$ is independent of all prior states $S_{\tau}$ for $\tau < t-1$ \cite{ghahramani}. The goal is to find the most likely sequence of hidden states. Generally speaking, an HMM is a time sequence of an observation sequence $X= \{ X_1, X_2, ..., X_n \}$, derived from a quantized codebook $V = \{ v_1, v_2, ..., v_{|V|} \}$, that is $X_k \in V, k =1, 2, ..., n$. In addition, hidden states $Y = \{ Y_1, Y_2, ..., Y_n \}$ are derived from the states in the system $S= {s_1, s_2, ..., s_{|S|}}$, that is $Y_k \in S, k =1, 2, ..., n$. 
The state transition matrix $A =\{ a_{ij} \} \; i,j=1...|S|$ models the probability of transitioning from state $s_i$ to $s_j$. Furthermore, $B= \{b_{jk} \} \; j =1...|S|, \; k = 1...|V| $ models the probability that the hidden state $s_j$ generates the observed output $v_k$. \\
\indent Figure~\ref{fig:HMM} shows an example of an HMM model with two hidden states, three observed states, and the corresponding state transition and output probabilities. This HMM is \emph{ergodic} because each hidden state can be reached from every other hidden states in one transition. A left-to-right HMM model does not have any backward transitions from the current state. 
We consider both ergodic and left-to-right HMM models. 
\section{Limitations of Existing Geture Recognition Approaches} \label{Challenges}
In theory, all machine learning algorithms tolerate noise in their
training data. Common approaches include using lots of training data,
adding features, and building better models, e.g., adding more
interior nodes to an HMM. In practice, we show that understanding
and measuring error inspires improvements in modeling and
classification.

\begin{figure}[!tb]
\centering
\includegraphics[height=1.2in, width=3.3in]{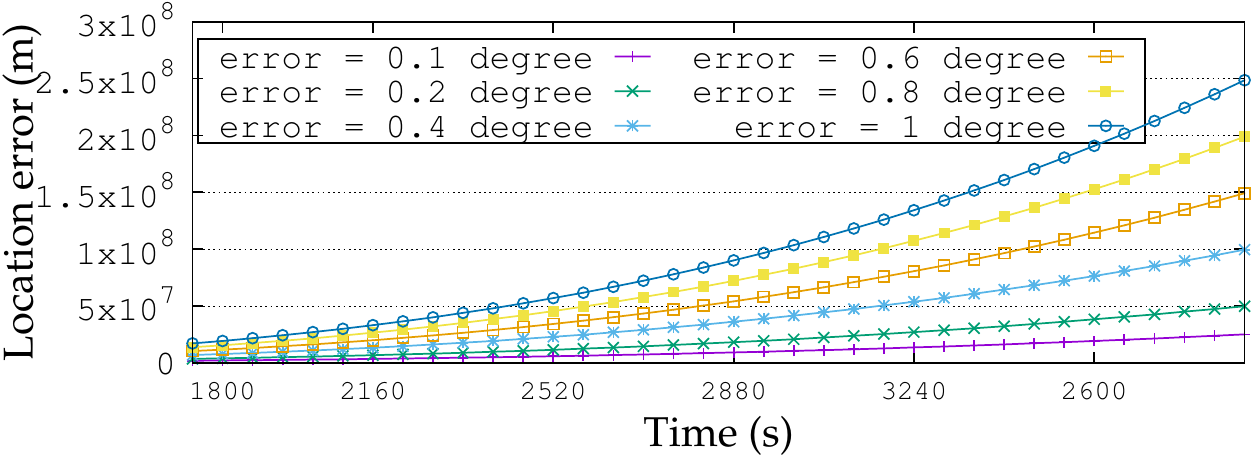} 
\caption{\label{fig:error} Location errors accumulate from various
  small angle
  errors.}
\end{figure}

\subsection{3-D Path Tracking is a Hard  Problem~\cite{woodman2007introduction}} A gesture is a meaningful 3-D
movement of a mobile device. When the phone moves, sensors gather an
observation consisting of a sequence of 3-D accelerometer data. We
use accelerometer observations to train 3-D gesture recognition
models, 
but our approach should apply to other sensors.

\begin{figure}[!tb]
\centering
\includegraphics[height=2in, width=2in]{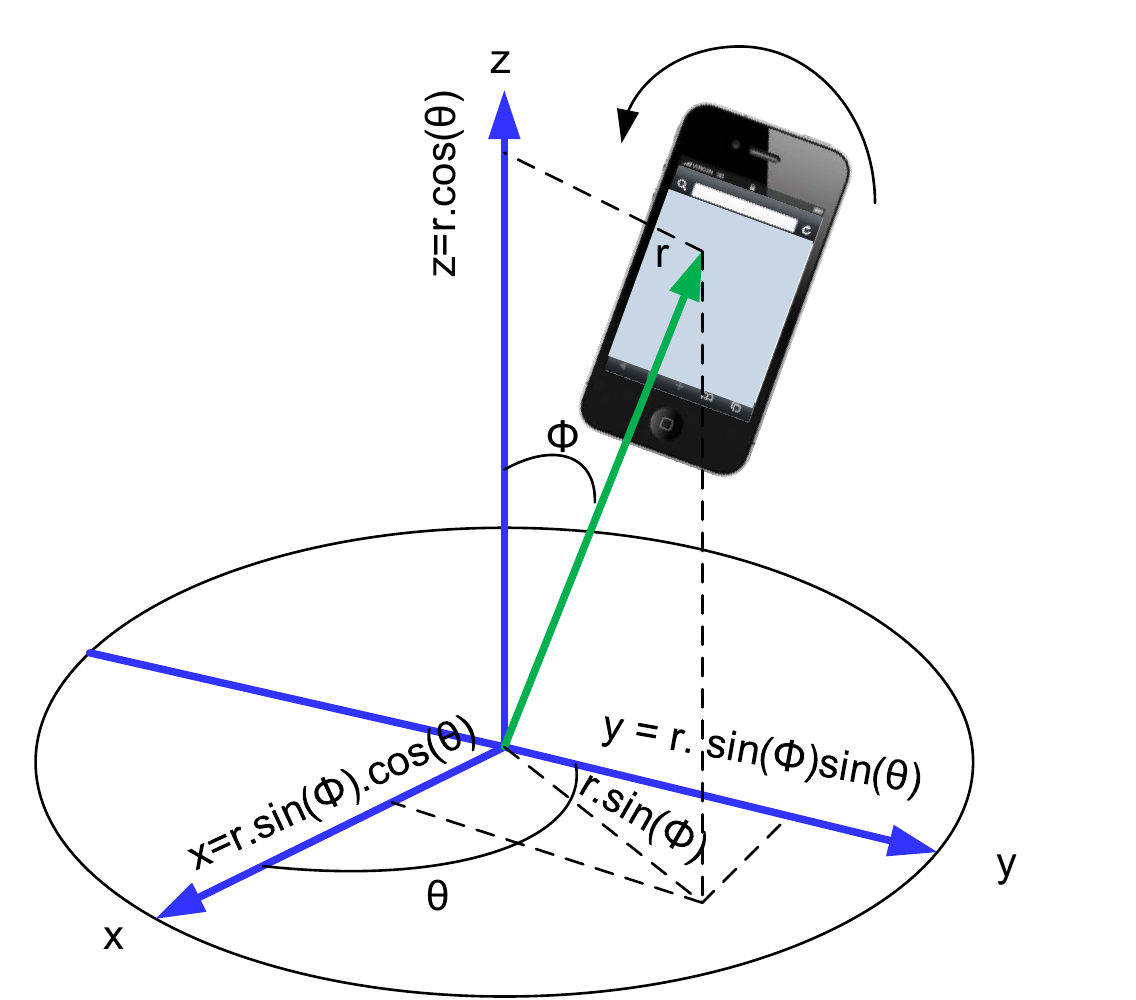} 
\caption{\label{fig:angle} The phone angle changes
  accelerometer readings.}
\end{figure}

Since users hold devices at different angles and move at different
velocities even when making the same gesture, the 3-D accelerometer
data includes a gravity component on all axes, which varies per user
and gesture. Figure~\ref{fig:angle} shows how the phone angle
generates different acceleration data projected on the X, Y, and Z
axes due to gravity. 
One approach we
tried was to eliminate gravity variation by using gyroscope data and tracking a 3-D path. 
This approach does not work because it actually amplifies error. We show this derivation to motivate the difficulty of path tracking and how gesture errors make it worse.

A 3-D path tracking approach models total acceleration and projects it to
a position as follows. Given
\begin{equation}
a_{\mathrm{m}}= a_{\mathrm{f}} - R(\alpha, \beta, \gamma) g \vec{z}
\end{equation}
where $a_{\mathrm{m}}$ is the measured data from the accelerometer;
$a_{\mathrm{f}}$ is the actual acceleration applied to the phone's
frame by the user; $R(\alpha, \beta, \gamma) = R_z (\alpha) R_y
(\beta) R_x (\gamma)$ is the rotation matrix between the actual force
applied to the phone and the frame of the sensor; and $\vec{z}$ is a
unique vector along z direction \cite{chatfield1997fundamentals,
  robotics:chrobotics}. Rotating the sensor frame acceleration to the
actual force frame gives the inertial acceleration:
\begin{equation}
a_{inertial}={R(\alpha, \beta, \gamma)}^{-1} a_{\mathrm{f}} = {R(\alpha, \beta, \gamma)}^{-1} a_{\mathrm{m}} + g \vec{z}
\end{equation}
Integration of the inertial acceleration produces a velocity and then
integrating acceleration twice produces a phone position.
\begin{equation}
    V(t) = \int a_{\mathrm{inertial}} dt 
\end{equation}
\begin{equation}
    R(t) = \int \int a_{\mathrm{inertial}} dt dt
\end{equation}

\noindent A rotation matrix is obtained by multiplying each of the yaw, roll, and pitch rotation matrices. Adding gyroscope data, and assuming the phone is still at $t=0$ (which means we know the initial angel with respect to gravity), the accumulated rotational velocity determines the 3-D angles with respect to gravity at any time \cite{chatfield1997fundamentals, robotics:chrobotics, foxlin2008motion}. Projecting the accelerator
data in this manner may seem appealing, but it is impractical for
two reasons. (1) It results in dimensionless gestures, which means the
classifier cannot differentiate a vertical circle from a
horizontal circle. Users would find this confusing.  (2) It amplifies
noise making machine learning harder. Figure~\ref{fig:error} shows the
accumulated error over time for different values of angle error. Even
small errors result in huge drift in tracking the location, 
making gesture tracking almost impossible. Consequently, we need to
use a different approach to handling gesture errors.

\begin{figure}[!tb]
\centering
\includegraphics[height=1.0in, width=3.3in]{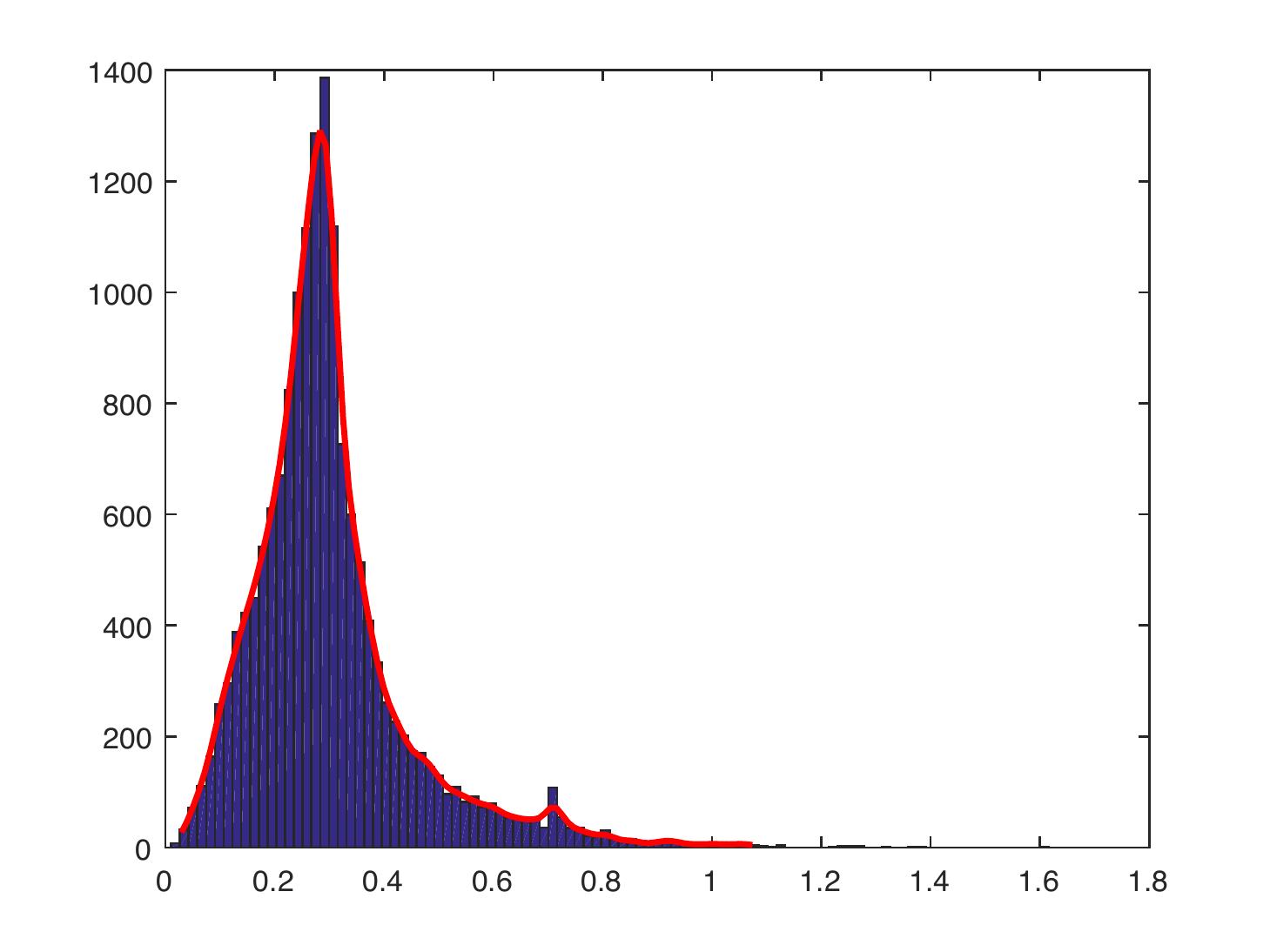}
\caption{\label{fig:ErrorDist-O} Error distribution of accelerometer
  data for ``O''.}
\includegraphics[height=1.0in, width=3.3in]{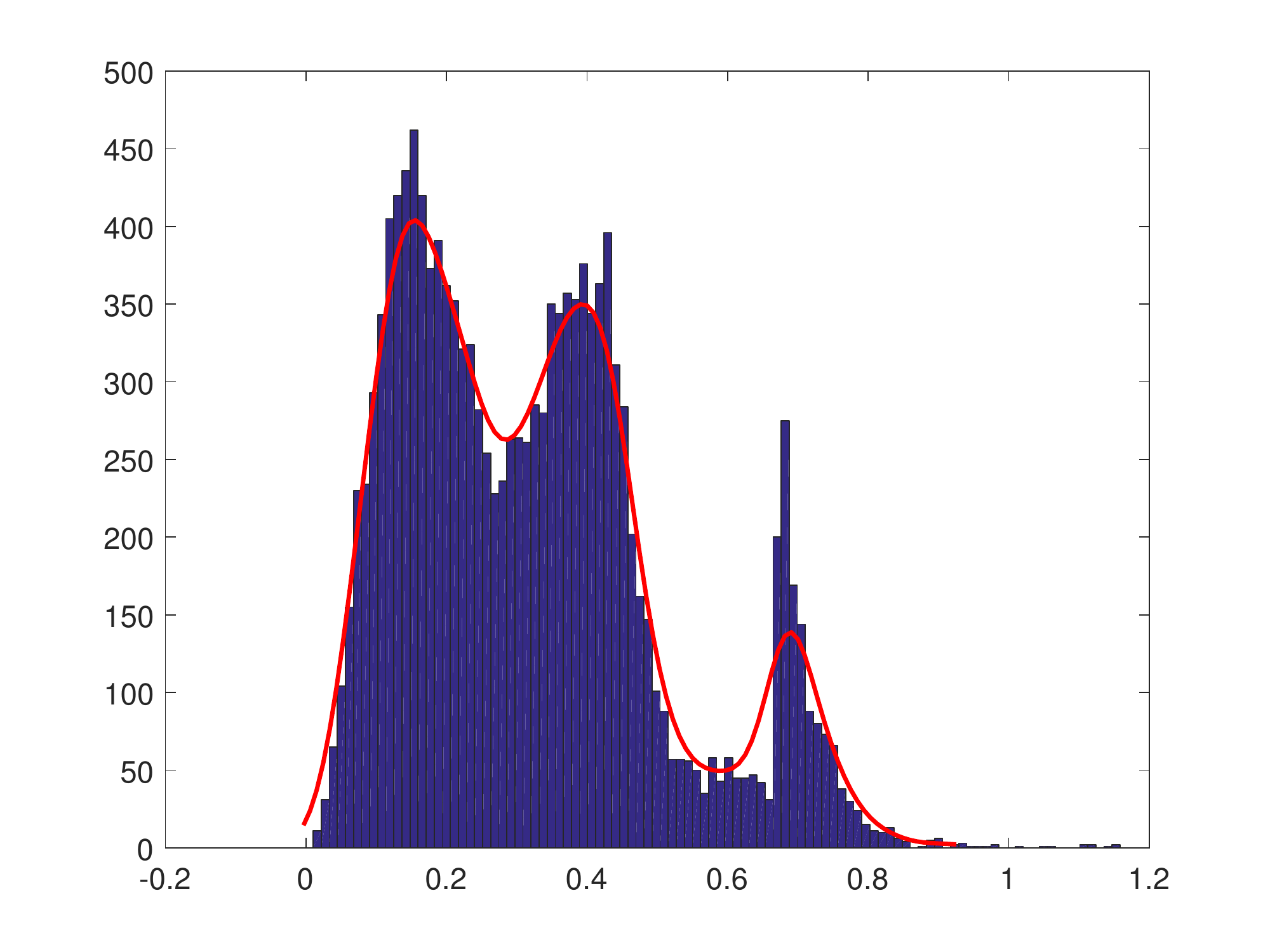}
\caption{\label{fig:ErrorDist-N} Error distribution of accelerometer
  data for ``N''.}
\end{figure}

\subsection{Noise is Gesture Specific}
We collect the error distribution, mean, and variance of the x, y, and
z accelerometer readings for each gesture at each position in sequence.
 Figures~\ref{fig:ErrorDist-O} and \ref{fig:ErrorDist-N} plot the
 resulting distributions for two examples from the ``O'' and ``N''
 gestures. The error distributions tend to be Gaussian mixture models
 since the accelerometer measure x, y, and z coordinates. Because the
 error is large and differs per gesture, it suggests different models
 for each gesture should be more accurate. 
The error distributions are not uniform. Mapping the data to codewords
can exploit this information. Prior approaches fall down on both fronts:
they do not learn separate models for each gesture or accomodate gesture noise in their codeword maps~\cite{ghahramani, lee, schlomer, zhang2009hand}

\subsection{Noise in Classification}
\begin{figure}[!tb]
\centering
\includegraphics[height=1.5in]{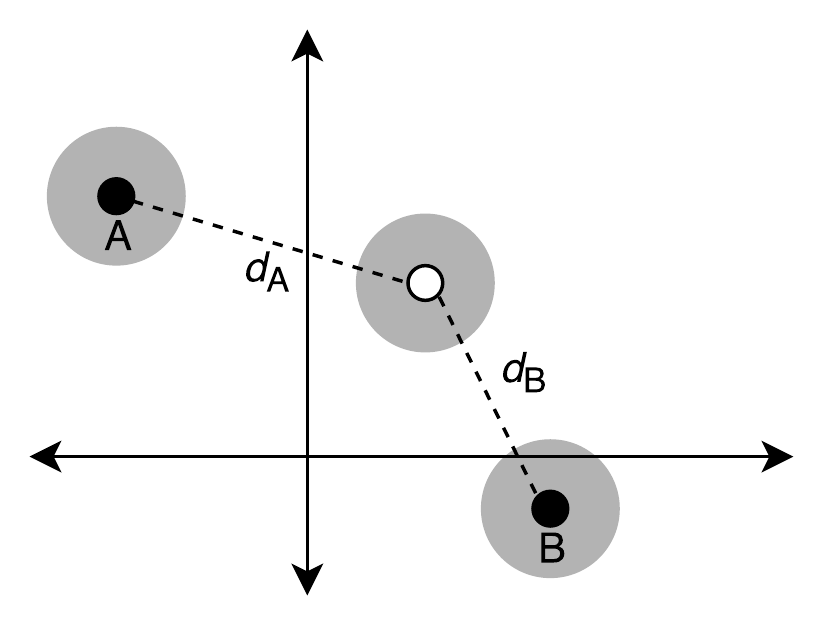}
\caption{\label{fig:quanterror} Error in sensor and codeword estimation
(shaded circles) is mitigated by \uncT.}
\end{figure}

Noise affects how the system maps a
sequence of continuous observations to discrete codewords. A deterministic quantization algorithm does not deal with this source of error.
Figure~\ref{fig:quanterror} illustrates why deterministic quantization is
insufficient. The black points are codewords and the white point is a sensed
position in 2D space. The distances $d_\mathrm{A}$ and $d_\mathrm{B}$ are similar, but the B codeword is slightly closer, so deterministic quantization would choose it. In reality, however, the sensed value is uncertain and so too are estimates of the codewords themselves. The gray discs show a confidence interval on the true position. For some values in the interval, $d_\mathrm{A} < d_\mathrm{B}$ and thus the correct quantization is A.
 
Our \emph{statistical quantization} approach explores multiple
codewords for the white point by assuming that points close to the
sensed value are likely to be the \emph{true} position. In other
words, the probability of a given point being correct is proportional
to its distance from the sensed value. We therefore choose the
codeword A with a probability proportional to $d_\mathrm{A}$ and B with a
probability proportional to $d_\mathrm{B}$.

\section{High Five: Gesture Training and Classification}\label{Approach}
\begin{figure}[!tb]
\centering
\includegraphics[height=2.3in, width=3.5in]{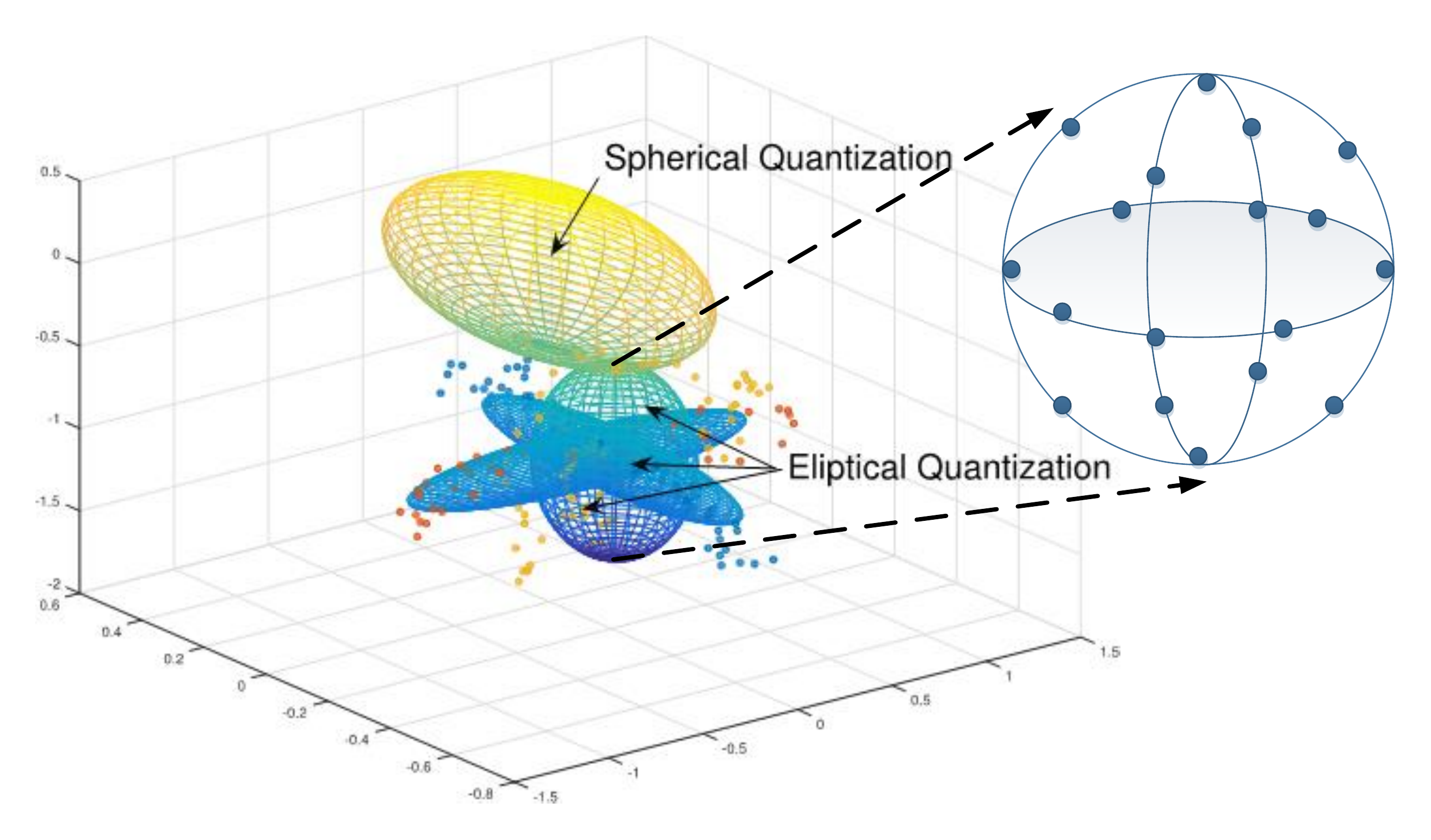}
\caption{\label{fig:Eliptical} Eliptical and spherical quantization
  for 3-D accelerometer data for ``x-dir'',  ``y-dir'', and  ``N'' gestures.}
\end{figure}

To accurately model and recognize gestures, we
propose two techniques: \emph{deterministic elliptical quantization} and
\emph{statistical elliptical quantization}.\\ 
\textbf{Deterministic Elliptical Quantization} During training, we
gather the statistical data on errors (distribution of error, mean,
and variance) for each position in gesture sequences,
create  codewords, and train HMMs for each gesture. We map all
the observation sequences for each gesture to a unique codebook. We
construct an elliptical contour for our codebook based on mean and
variance of the observations. Figure~\ref{fig:Eliptical} shows the spherical
equal-spaced codebook generated for all gestures based on prior work~\cite{schlomer} and our per-gesture
ellipses for three example gestures.  In the figure,
the acceleration data is expressed in terms of $g \simeq 9.8$. If we hold the phone along the Z axis, the scattered data has a bias of $(0, 0, -1)$ which shows the gravity component. If the user holds the phone upside-down the scattered data has a bias of $(0, 0, 1)$ and our statistical generated codewords embraces the gravity component in each case.
Per-gesture ellipses better fit the data than a single sphere for all gestures. We use 18 equally spaced points on the elliptical contour to ease comparison with related work, which uses 18 points on a spherical contour~\cite{schlomer}. 18 observation states strikes a balance between learning complexity and accuracy, both of
which are a function of the number of states. This method is similar
to multi-dimensional data scaling~\cite{kruskal1978multidimensional}, but as we showed in the previous section, standard projection is a poor choice for this data. We construct elliptical models for each gesture as follows.
\begin{equation}
\frac{(x - \mu_x)^2}{( \sigma_x)^2} + \frac{(y - \mu_y)^2}{( \sigma_y)^2} + \frac{(z - \mu_z)^2}{(\sigma_z)^2}  = 1
\end{equation}
The values $\mu_x$, $\mu_y$ and $\mu_z$ are the expected value of raw
acceleration data for each gesture. We construct a different codebook
for each gesture.
This process maps the accelerometer data to one of the 18 data points as shown in Figure~\ref{fig:Eliptical}. The mapped data constructs the observed information in the Hidden Markov Model.

Our quantization approach differs from the prior work~\cite{schlomer} in two ways. 
First, since we use the statistics of each gesture, there is no need
to remove the gravity bias, because the center of mass for all gesture
data of a specific gesture includes the gravity component.  The second  difference
is that we chose a different contour for each gesture in our data
set. As Figure~\ref{fig:Eliptical} shows, the elliptical contour for a
\textbf{x-dir} gesture is completely different from the contour for
\textbf{y-dir} or \textbf{N}.
In the spherical contour, most of the data points from the
accelerometer map to a single codeword, 
eliminating a lot of information that is useful for
classification. Our approach reduces the quantization error for
different gestures since it is much less likely to map
each gesture to another gesture's codebook and generate the same
sequence.\\

\textbf{Gesture Training}
After measuring the noise and training the codebooks for each gesture, we build an HMM model for each gesture. 
The gesture recognition application takes as input the 3-D
accelerometer sequence for each gesture and updates the HMM probabilities using the forward-backward algorithm \cite{baum,
  schlomer, lee, mitra}. We use Baum--Welch filters to
find the unknown parameters of each gesture's HMM model (i.e., $a_{ij}$ and $b_{jk}$) \cite{baumLE}. Assuming $P(X_t | X_{t-1})$ is independent of time $t$ and assuming the probabilities of initial states is $\pi_i = P(X_1 = i)$, the probability of a certain observation at time $t$ for state $j$ is given by
\begin{equation}
b_j(y_t) = P(Y_t =y_t | X_t =j)
\end{equation}
Baum--Welch filters use a set of Expectation Maximization (EM) steps. Assuming a random initial condition $\theta = (A, B, \pi)$ for the HMM, Baum--Welch finds the local maximization state transition
probabilities, output probabilities, and state probabilities. That is,
HMM parameters $\theta^{\ast}$ which will maximize the observation
probabilities as follows.
\begin{equation}
\theta^{\ast} = \mathrm{argmax}_{\theta} {P(Y|\theta)}
\end{equation}
The algorithm iteratively updates $A$,
$B$, and $\pi$ to produce a new HMM with a higher probability of
generating the observed sequence. It repeats the update procedure
until it finds a local maximum. In our deployment, we store one final HMM for each gesture as a binary file on the phone.\\ 

\textbf{Statistical Gaussian Mixture Model (GMM) Quantization} 
The key to our \emph{Statistical Elliptical Quantization} approach is
representing each gesture reading as a random variable that
incorporates its sensor noise. For classification,  our
statistical quantization approach uses the Gaussian distribution
mixture models based on the error model we observe for each gesture
during training. For example, Figures~\ref{fig:ErrorDist-O}
and~\ref{fig:ErrorDist-N} show that the probability distribution of distance of $data$ mapped to $code_i$ follows a Gaussian mixture model distribution with 3 peaks each representing the peak over one of the X, Y or Z coordinate. The probability of mapping a data point to each codeword for a bivariate Gaussian noise distribution is computed as follows:
\begin{equation}\scriptsize
P(code_i) = \frac{\frac{1}{\sqrt{2 \pi \sigma_{i,x}^2 \sigma_{i,y}^2 \sigma_{i,z}^2}}e^{\sum_{k= x,y,z} \frac{-{(d_{i,k}- \mu_{i,k})}^2}{2 \sigma_{i,k}^2}  }}{\sum_{j=1}^{N} \frac{1}{\sqrt{2 \pi \sigma_{j,x}^2 \sigma_{j,y}^2 \sigma_{j,z}^2}}e^{\sum_{k= x,y,z} \frac{-{(d_{j,k}- \mu_{j,k})}^2}{2 \sigma_{j,k}^2} } }
\end{equation}
A mixture of three Gaussian distribution models maps to individual Gaussian models as follows:
\begin{lstlisting}[escapeinside={/*@}{@*/}]
/* Quantization code for a mixture of three Gaussian distribution. */
var acc = ReadAccelerometer();
var d = /*@ $\|{(acc, codeword_i)}\|$ @*/ ;
if /*@ $ d < ( \mu_1 + \mu_2)/2$ @*/ : 
/*@ $d = N( \mu_1, \sigma_1 )$ @*/ ;
else if /*@ $ ( \mu_1 + \mu_2)/2 < d < ( \mu_2 + \mu_3)/2 $ @*/ :
/*@ $ d = N( \mu_2, \sigma_2)$ @*/ ;
else: 
/*@ $d = N( \mu_3, \sigma_3)$ @*/ 
\end{lstlisting}
\noindent This mapping produces a probability distribution over codewords for each
reading. Sampling from this distribution creates multiple sequences
of observation for the HMM, which then
determines the most likely gesture from the entire distribution.

\textbf{Statistical Random Quantization}
For comparison, we also implement a random quantizer to exploit error
if training did not or was not able to produce a gesture-specific error model.
This quantizer maps an observation \emph{randomly} to codewords depending on their distance to the data point. 
For example, given four codewords, it randomly maps the gesture data proportional to the distance with respect to each codeword:
\begin{equation}
    P(\mathrm{code}_i) = \frac{\frac{1}{d_i}}{\sum_{j=1}^{N} \frac{1}{d_j}}
\end{equation}
%

%

\textbf{Gesture Classification}
The GMM quantization and Random quantization algorithms appear
in Algorithms\ref{RandomClassifier} and~\ref{NaiveClassifier}, respectively.
We implement these classifiers in the \uncT programming language
(described below),
exploiting its first-class support for
probability distributions. 

Algorithm~\ref{RandomClassifier} shows our novel
statistical RMM quantization. Each step of the algorithm maps user data to a sequence of observation states from the generated codebook during training for each of $N$ gestures in $G = \{G_1, G_2, ... G_N\}$. We treat the mapping independently for each data point in $G_i$. (We also explored computing correlated mapping where mapping the current 3-D data to one of the quantization codewords depends on the previous 3-D mapping, which further improves accuracy, but for brevity omit it). At each
step, we sample nearby codewords in $G_i$ and weigh them by their
probability based on the GMM error model observed during training to create a sequence of observation states. We next classify the generated sequence to find $P(G_i|X)$, sample until the probabilities converge,  and then pick the most likely sequence.  When the algorithm completes, we have computed the most likely HMM path for each $G_i$. We only consider $G_i$ with a probability above a threshold $thr$ as potential gestures and thus may not return a
gesture. For those $G_i$ above the threshold, we return the one with
the highest probability as the most likely gesture. We explore $thr$
values of 0.5 and $1/N$ and find that 0.5 works best.\\

Algorithm~\ref{NaiveClassifier} uses the Random quantizer which
implements a Bayesian classification scheme that returns the
probability of each gesture given the observed sequence
\cite{russell}. Given a set of observation sequence $X= \{ X_1, X_2,
..., X_n \}$, it computes the probability of each gesture $G_k$ as follows.
\begin{equation}
P(G_k | X) = \frac{P(G_k) P(X | G_k)}{P(X)}
\end{equation}
The values $P(X | G_k)$ and $P(G_k)$ are produced by the Baum--Welch
training model for each individual gesture. 

\begin{algorithm}[!tb]
 \KwData{Raw accelerometer data, HMM models for $G_i \in G$, Inference threshold ($thr$)}
 \KwResult{If $(gesture == G_i).Pr \geq thr$ or $(gesture != G_i).Pr \geq thr$}
  Most probable gesture =$\{\}$ \;
 \For{$G_i \in G$ } {
 $(gesture == G_i).Pr = 0 $\;
\While{$(gesture == G_i).Pr \le thr$ or $(gesture != G_i).Pr \le thr$}{
  X= Map Accelerometer data sequence to to $G_i$'s quantization codebook using random quantization\;
  Find $P(X|G_i)$ from the Baum--Welch trained HMM model\;
  $P(G_i | X)= \frac{P(G_i) P(X | G_i)}{P(X)} $\;
  Add $P(G_i | X)$ to $G_i$'s distribution model\;
  }
  \If{$(gesture == G_i).Pr \geq thr$ and $(gesture == G_i).Pr \geq MaxProb$}{
  MaxProb = $(gesture == G_i).Pr$ \;
  Most probable gesture = $G_i$ \;
  }
  }
  return Most probable gesture\; 
 \caption{High Five: GMM Quantization}
 \label{RandomClassifier}
\end{algorithm}

\begin{algorithm}[!tb]
 \KwData{Raw accelerometer data, HMM models for $G_i \in G$}
 \KwResult{The most probable gesture}
 MaxProb = 0 \;
 Most probable gesture =$\{\}$ \;
 \For{$G_i \in G$ } {
  Map Accelerometer data to to the quantization codebook for each gesture using deterministic quantization\;
  Find $P(X|G_i)$ from the Baum--Welch trained HMM model\;
  $P(G_i | X)= \frac{P(G_i) P(X | G_i)}{P(X)} $ \;
  \If {$ P(G_i | X) \geq MaxProb$ }{
  MaxProb = $P(G_i | X)$ \;
  Most probable gesture = $G_i$ \;
  }
  }
  return Most probable gesture\;
 \caption{High Five: Random Quantization}
 \label{NaiveClassifier}
\end{algorithm}
\section{Discussion}
Our HMM model tracks a single HMM path during classification but builds many possible input observations from a single trace of accelerometer data.  In contrast, prior work shows it is possible to use an HMM which tracks the $k$ top paths during 
classification~\cite{seshadri1994list}. It is interesting future work to explore any experimental differences in such a formulation.  We use the raw
accelerometer data as a feature given to our HMM training and classification.
However, there exist prior works, especially in computer vision, that finds rotation-invariant or scale-invariant features \cite{lowe1999object, lowe2004distinctive}. We did not use rotation-invariant features because we want the capability to define more gestures, e.g., ``N'' in the x-y and in the z-y plane are distinct. However, more sophisticated features can further improve our classification accuracy.
\section{Implementation}
To help developers create models for problems in big data, cryptography, and artificial intelligence, which benefit from probabilistic reasoning, researchers have recently proposed a variety of probabilistic programming languages \cite{bornholt2014uncertain,mitchell2006probabilistic, xie2010using, ngo1997answering, kiselyov2009embedded,sampson2014expressing}.
We use the \uncT programming language to
implement random quantization. We choose it, because \uncT is sufficiently expressive and automates inference, and thus significantly simplifies our implementation~\cite{bornholt2014uncertain, sampson2014expressing, nandi2017debugging}. 
The remainder of this section gives background on \uncT, the programming model that
inspired and supports our technique, and describes our implementation.

\subsection{The Uncertain$\langle$T$\rangle$ Programming Model}
\uncT is a generic type and an associated runtime in which developers
(i) express how uncertainty flows through their computations and (ii) how to act
on any resulting uncertain computations. To accomplish (i) a developer annotates
some type $T$ as being uncertain and then defines what it means to sample from
the distribution over $T$ through a simple set of APIs.  Consumers of this type
compute on the base type as usual or use LINQ primitives~\cite{Meijer} to build derived computations. The
\uncT runtime turns these derived computations into a distribution over those
computations when the program queries it.  Querying a distributions for its
expected values and or executing a hypothesis test for conditionals, triggers a statistical test.
Both of these queries free the \uncT runtime from exactly representing a
distribution and let it rely on lazy evaluation 
and
sampling to ultimately determine the result of any query.

\subsection{Statistical Quantization with Uncertain$\langle$T$\rangle$}
\label{sec:rqimpl}

\begin{scriptsize}
\begin{figure}[!tb]
\begin{lstlisting}
/* Classification Code for Statistical Quantization. */
var acc = ReadAccelerometer();
Uncertain<int> gestures = 
  // distribution over observations
  from obs in new StatisticalQuantizer(acc)
  // returns most likely gesture
  let gesture = Bayes.Classify(acc, obs) 
  select gesture;

// T-test: more likely than not that
// this is the gesture labeled 0
if ((gestures == 0).Pr(0.5))
	Console.WriteLine("gesture=N");
\end{lstlisting}
  \caption{Statistical Quantization for a single gesture.}
  \label{fig:MyStatisticalQuantizer}
\end{figure}
\end{scriptsize}
To implement statistical quantization, we express
each gesture as a random variable over integer labels. Our implementation of \lstinline{StatisticalQuantizer(acc)} (Figure~\ref{fig:MyStatisticalQuantizer}) first reads from the accelerometer and passes this observation to the
\lstinline{RandomQuantizer(acc)} constructor which knows how to sample from
observations by randomly mapping analog accelerometer data to discrete codewords
to return a distribution over observations. The LINQ syntax let's the developer
call existing code designed to operate on type $T$ (i.e.,
\lstinline{Bayes.Classify} which operates on concrete observations) and further
lets her describe how to lift such computation to operate over distributions.
In gesture recognition, the resulting type of \lstinline{gestures} is then an \lstinline{Uncertain<int>}
or a distribution over gesture labels.

The \uncT runtime does not execute the lifted
computations until the program queries a distribution's \emph{expected value}
or uses it in a conditional test. For example, when the developer writes
\lstinline{if ((gestures == 0).Pr(0.5))} the \uncT runtime executes a hypothesis
test to evaluate whether there is enough evidence to statistically ascertain
whether it is more likely than not that the random variable \lstinline{gestures}
is equal to the gesture with label 0.  The runtime samples from the leaves of
the 
program and propagates concrete values through any user-defined
computation 
until enough evidence is ascertained to
determine the outcome of a conditional.  The \uncT runtime implements many
inference algorithms under the hood (rejection sampling, Markov Chain Monte
Carlo, etc.).  For this domain, we found no reason to prefer one over the other
and so use rejection sampling for all experiments.

\begin{figure}[!tb]
\centering
\includegraphics[height=1.5in, width=2.5in]{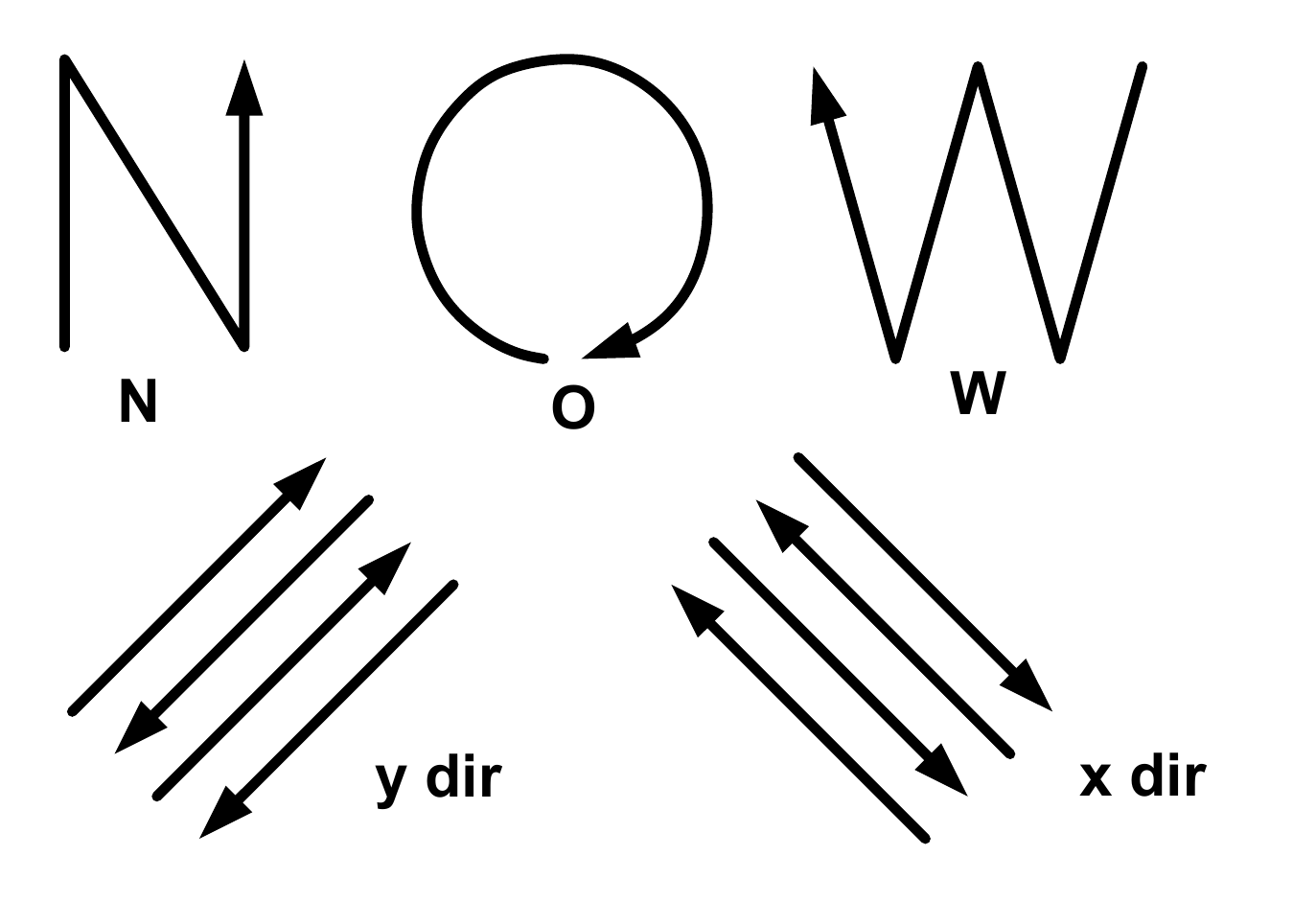}
\caption{\label{fig:Gestures}Five gestures in Windows Phone (WP) data set.}
\end{figure}
\begin{figure}[!tb]
\centering
\includegraphics[height=3.2in, width=3.2in]{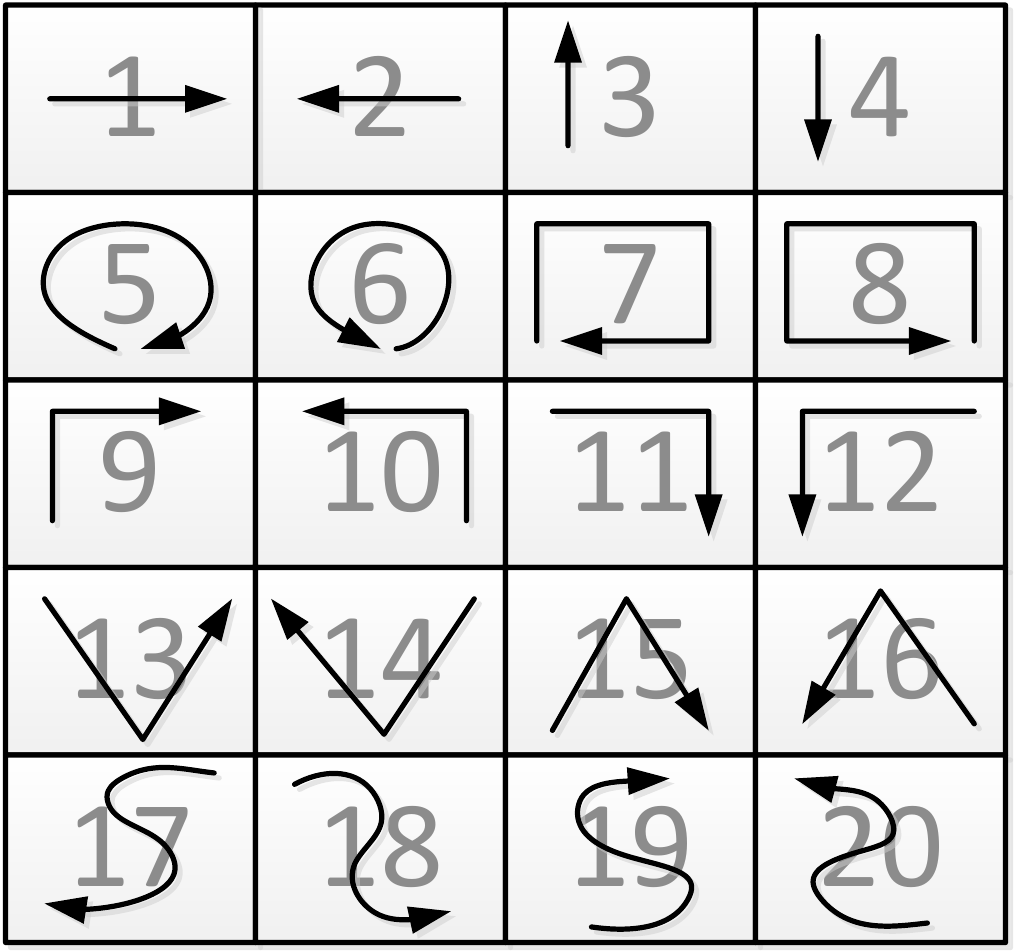}
\caption{\label{fig:dataset2}Twenty gestures in Smart-Watch (SW) data set \cite{costante2014personalizing}.}
\end{figure}

\section{Methodology} \label{Evaluation Methodology}

This section describes the data sets, implementation details, and
algorithms that we use to evaluate statistical quantization. We
evaluate our algorithms and collect one data set on a smart-phone
Nokia Lumia 920 (Windows Phone 8). We use the Windows SDK 8 API to
read the 3-D accelerometer.

\textbf{Data sets, training, and testing} We collect our own data set on the Windows Phone and use the publicly available Smart Watch data set from Constante et~al.~\cite{costante2014personalizing}, which together make a total of 4200 gesture samples of 25 distinct gestures performed by 28 people.

\begin{description}
\setlength{\itemsep}{-0ex}
\item[Windows Phone (WP) Data Set] We collect data from 10 men and 10 women performing 10 times each of the 5 gestures shown in Figure~\ref{fig:Gestures} on our Windows Phone platform, for a total of 1000 gesture samples. 
\item[Smart Watch (SW) Data Set] We also use a publicly available data set consisting of the 20 gestures  shown in Figure~\ref{fig:dataset2} trained by eight people~\cite{costante2014personalizing}. Each person performed each gesture 20 times for a total of 3200 samples. 
\end{description}
Prior studies~\cite{schlomer, liu2009uwave} have smaller data sets and very distinct gesture patterns. In contrast, our data sets include gestures with very similar patterns. For example, W and N in the WP data set differ by about one stroke, and G9 and G11 differ by a 90$^{\circ}$ rotation, making them hard to differentiate.

These data sets represent a realistic amount of training for individuals, because users, even paid ones, are unlikely to perform the same gesture well 100s of times for training. Training must be short and recognition must be effective quickly to deliver a good user experience. To create sufficient training data, we train each classifier with data from all the users (20 for WP and 8 for SW).  
To assess the accuracy of the gesture recognition algorithms, we randomly split the data sets into 75\% training data and  25\% test data and repeat this procedure 10 times.

\textbf{The High Five gesture recognition application} We implement a Windows Phone gesture recognition application, called \emph{High Five}.  Users train the system online by first specifying a new gesture and then train the system by performing the gesture at least 10 times. Users can perform any gesture they want and then specify a corresponding action triggered by the gesture (e.g., call Mom on M, send an Email on E). The application has two modes: \emph{signaled}, in which users open the gesture recognition application first before making a gesture, and \emph{dead start}, which captures all device movements, and thus is more likely than signaled recognition to observe actions that are not gestures.  
We implement the system in Microsoft's Visual Studio 2015 for C\# and the Window's Phone software development kit (SDK) sensor API. 
 We use the \uncT libraries and runtime for our statistical quantizer by adding an HMM API to \uncT that returns samples from HMM distributions. 

\textbf{Gesture recognition algorithms} We evaluate the following gesture recognition algorithms.

\begin{description}
\setlength{\itemsep}{-0ex}
\item[Deterministic Spherical Quantizer] Wijee is the prior state-of-the-art~\cite{schlomer}. It uses a traditional left-to-right HMM with kmeans quantization and one spherical model for all gestures~\cite{schlomer}. We follow their work by limiting transitions to four possible next codewords, such that $S_i$ is the only possible next state from $S_j$ where $i \leq j \leq i+3$.
(They find that left-to-right and ergodic HMMs produce the same results.) We extend their algorithm to train gesture-specific models using a unique codebook for each gesture. 
Since the scattered data is different for each gesture, using per-gesture codebooks offers substantial improvements over a single codebook for all gestures.

 
\item[Deterministic Elliptical Quantizer] This algorithm uses a left-to-right HMM, elliptical quantization, and a unique codebook for each gesture.

\item[Statistical GMM Quantizer] This algorithm uses a left-to-right HMM, statistical Gaussian mixture model (GMM) elliptical quantization based on observed error, and a unique codebook for each gesture. The runtime generates multiple observation sequences by mapping the data sequences to multiple codeword sequences for each gesture using a gaussian mixture model. With statistical quantization, the developer chooses a threshold that controls false positives and negatives, which we explore below.

\item[Statistical Random Quantizer] This algorithm uses a left-to-right per-gesture elliptical HMM, statistical random quantization, and a unique codebook for each gesture.
\end{description}

\begin{figure*} [!tb]
        \centering
        \begin{subfigure}[b]{0.32\textwidth} 
                \includegraphics[height=1.5in, width=2.4in]{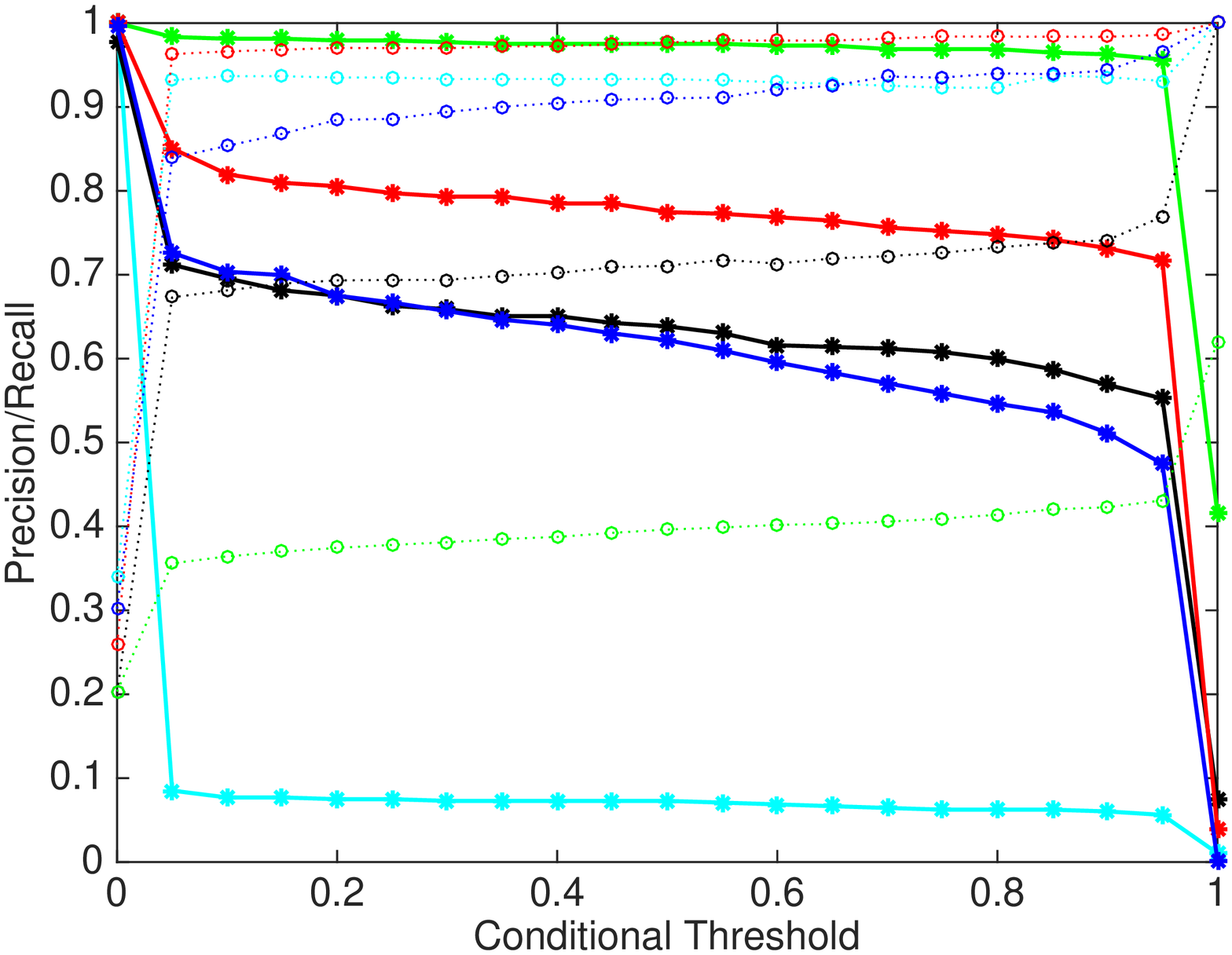}
                \caption{\label{fig:wijee}Deterministic spherical quantizer}
        \end{subfigure}%
        ~ 
        \begin{subfigure}[b]{0.32\textwidth}
                \includegraphics[height=1.5in, width=2.4in]{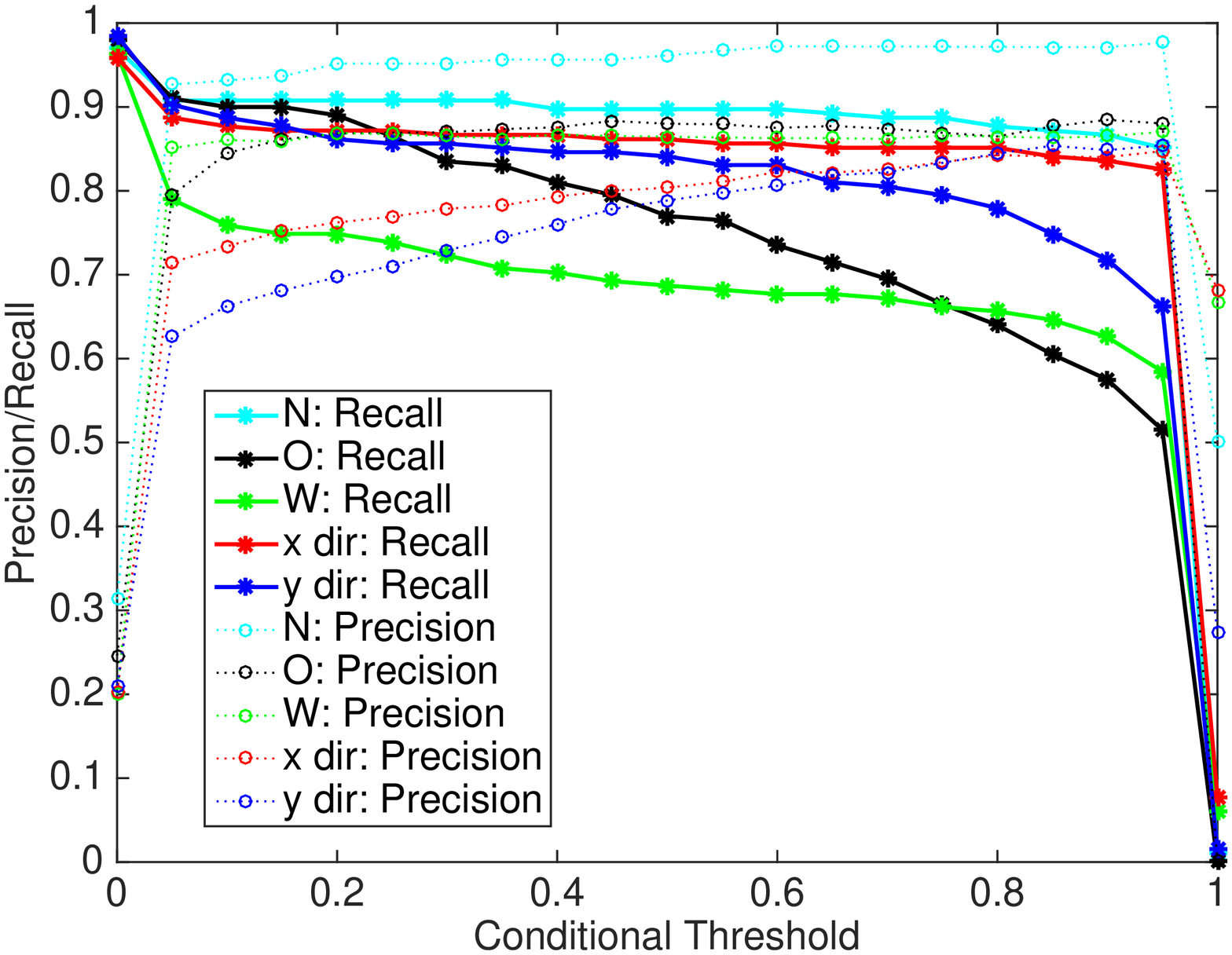}
                \caption{\label{fig:hi5}Deterministic elliptical quantizer}
        \end{subfigure}
        ~ 
        \begin{subfigure}[b]{0.32\textwidth}
                \includegraphics[height=1.5in, width=2.4in]{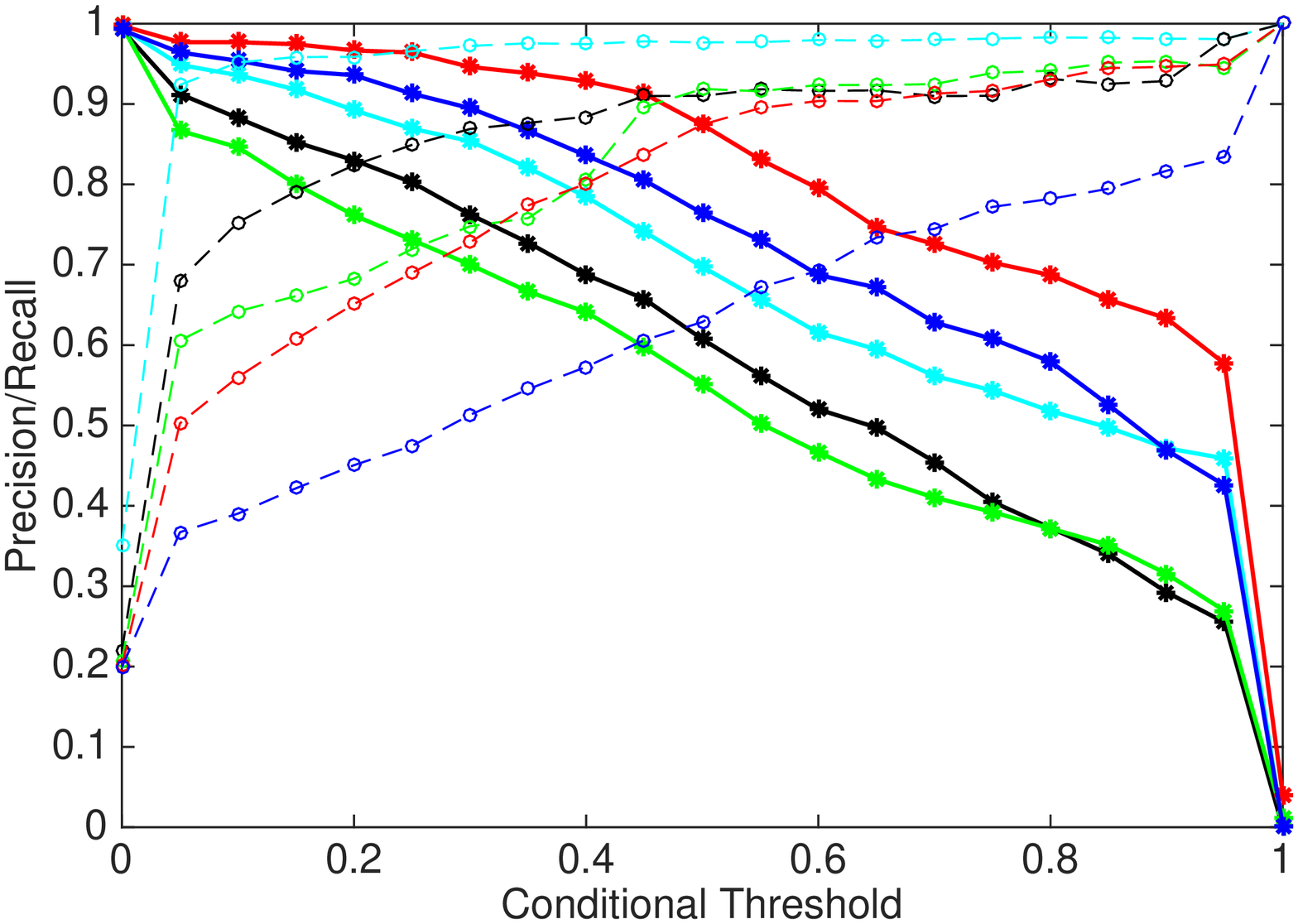} 
                \caption{\label{fig:random}Statistical GMM quantizer}
        \end{subfigure}        
        \caption{\label{fig:P_R}Precision and recall curves for gesture recognition algorithms.}
\end{figure*}

\section{Evaluation} \label{experimental}
This section compares the precision, recall, and recognition rate of the gesture recognition algorithms. We show that statistical quantization is highly configurable and offers substantial improvements in accuracy, recall, and recognition over other algorithms. The other recognizers are all much less configurable and achieve lower maximum accuracy, recall, and/or recognition in their best configurations.  These experiments illustrate that a key contribution of statistical quantization is that it has the power to offer both (1) highly accurate recognition in the signaled scenario, and (2) significant reductions in false positives in the dead-start scenario, thus matching a wide range of application needs.

We explore the sensitivity of gesture classification accuracy as a function of the number of gestures, using 2 to 20 SW gestures. For all the algorithms, accuracy improves with fewer gestures to differentiate. Statistical random quantization is however substantially more accurate than the others for all numbers of gestures. We further show that our approach is relatively insensitive to the number of users in the training data. Finally, we show how to easily incorporate  personalization based on other factors, such as performing a subset of the gestures, and the result further improves accuracy.

\textbf{Recognition rates for signaled gestures} In this first experiment, users open the High Five application and then perform the gesture, signalling their intent. 
Figure~\ref{fig:P_R} shows precision (dashed lines) and recall (solid lines) for each of the 5 gestures in distinct colors for the WP data set as  a function of the conditional threshold. Precision is the probability that a gesture is correctly recognized and is an indication of false positives while recall shows the probability of recognizing a performed gesture and shows false negatives. The deterministic elliptical quantizer in Figure~\ref{fig:P_R}(b) uses the domain specific knowledge of each gesture during training and thus has higher precision and recall compared to deterministic spherical quantization in Figure~\ref{fig:P_R}(a).  

Statistical GMM quantization in Figure~\ref{fig:P_R}(c) offers further improvements in precision and recall. Although the recall curve goes down as a function of the conditional threshold, when the conditional threshold is $1/N$ or lower, the recognition rate is higher than deterministic elliptical quantization. Statistical GMM delivers a similar threshold for precision. Statistical GMM quantization offers a distinct and smooth trade-off between precision and recall. Applications thus have a range of choices for the conditional threshold from which to choose that they can tailor to their requirements, or even let users configure. For instance, when the user is on a bus, she could specify higher precision to avoid false positives, since she does not want the phone to call her boss with an unusual movement of the bus. Prior work requires the training algorithm specify this tradeoff, instead of the end developers and users. 

\begin{figure}[!tb]
\centering
\includegraphics[height=1in, width=3.4in]{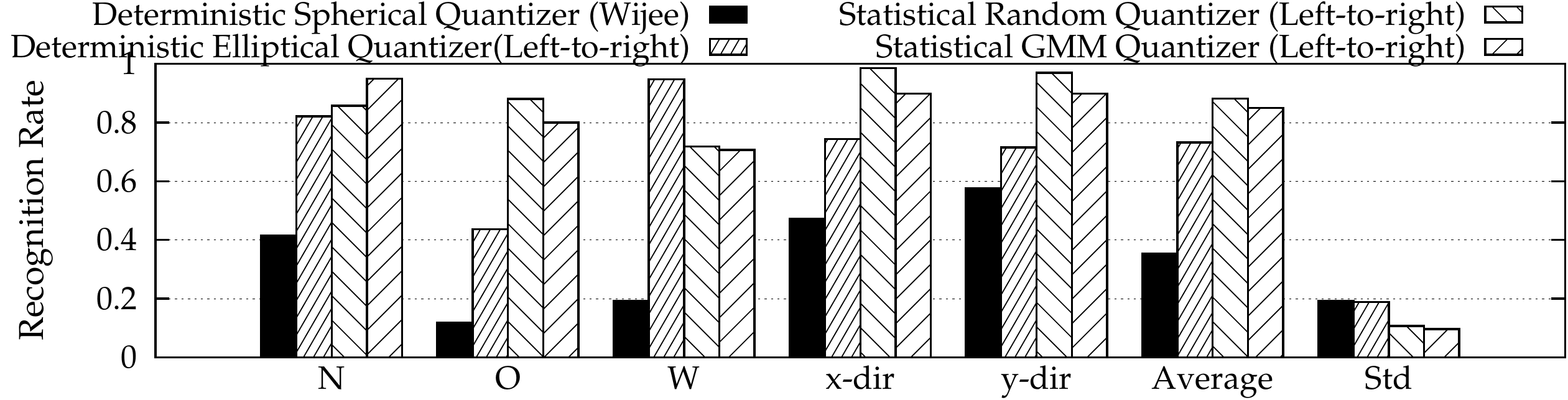}
\caption{\label{fig:RecognitionRate} Gesture recognition rates for WP data set.}
\end{figure}

Figure~\ref{fig:RecognitionRate} shows the recognition rate for each gesture in the WP data set for all the classifiers. The deterministic elliptical quantizer improves substantially over the deterministic spherical quantizer. Statistical GMM and random quantization deliver an additional boost in the recognition rate. Both GMM and random produce similar results within the standard deviation plotted in the last columns. On average, both statistical GMM and random quantization deliver a recognition rate of 85 and 88\%, respectively, almost a factor of two improvement over deterministic spherical quantization.  

\textbf{Recognition rates for dead start and as a function of gestures} This experiment explores the ability of the gesture recognition algorithm to differentiate between no gesture and a gesture since users do not signal they will perform a gesture. For instance, a gesture M must both wake up the phone and call your mom. In this scenario, controlling  false positives when you carry your phone in your pocket or purse is more important than recall---you do not want to call your mom unintentionally.



\begin{figure}[!tb]
\centering
\includegraphics[height=1in, width=3.4in]{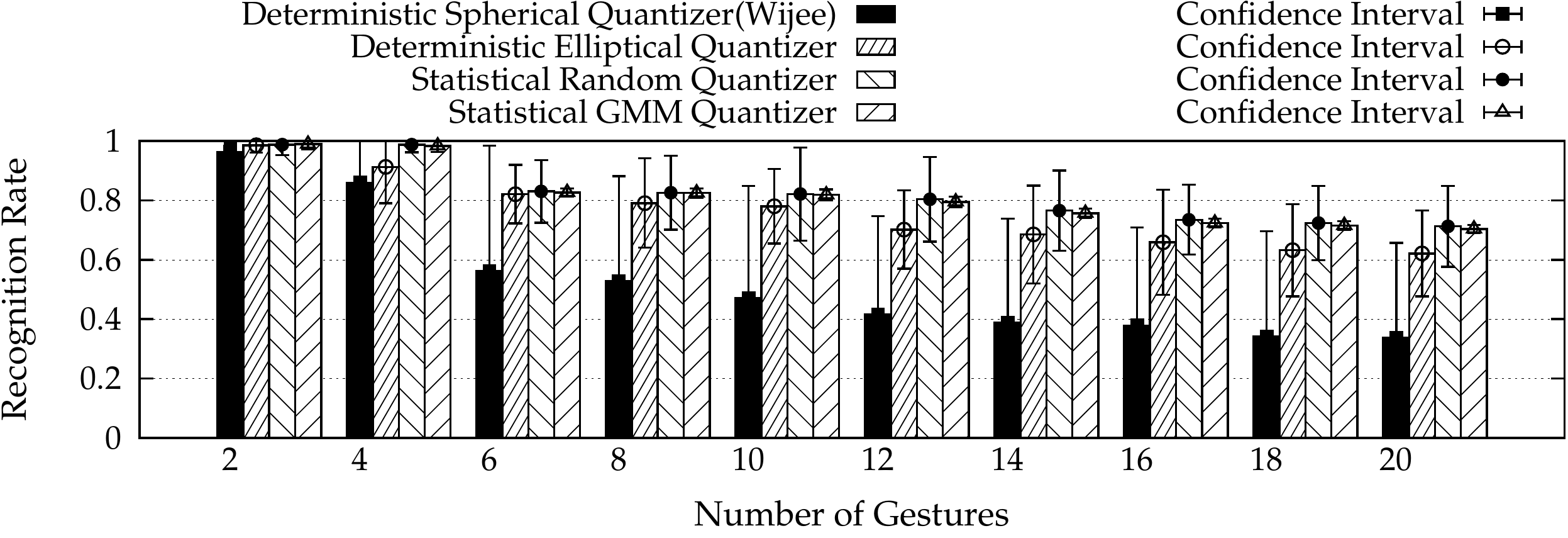}
\caption{\label{fig:IncreasingGestures} Gesture recognition rate as a function
of the number of gestures in the SW data set. 
}
\end{figure}

\textbf{Accuracy as a function of the number of gestures}  
The more gestures the harder it is to differentiate them. To explore this sensitivity,  we vary the number of gestures from 2 to 20 and compare the four approaches. Figure~\ref{fig:IncreasingGestures} shows the recognition rate for the deterministic spherical, deterministic elliptical, statistical random, and statistical GMM quantizers as a function of the number of gestures in the High Five application. 
All classifiers are more accurate with fewer gestures compared to more gestures.  Increases in the number of gestures degrades the recognition rate of the deterministic spherical faster than compared to the other classifiers.  Both deterministic spherical and elliptical classification have high variance. The statistical quantizers always achieves the best recognition rates, but GMM has a lot less variance than random, as expected since it  models the actual error. For instance, GMM achieves a 71\% recognition rate for 20 gestures, whereas deterministic spherical quantizer has a recognition rate of 33.8\%. Statistical GMM quantization has a  98\% recognition rate for 2 gestures.

\textbf{User-dependent and user-independent gestures} To explore the sensitivity of recognition to the training data, we vary the number of users in the training data from 2 to 8.  We compare with  Costante et~al.~\cite{costante2014personalizing} and Liu et~al.~\cite{liu2009uwave} which both perform this same experiment. We use six gestures from the Costante et~al.\/ SW data set: gestures 1, 3, 5, 7, 9, and 11.  Costante et~al. find that more users produces better accuracy, whereas Liu et~al. find more personalized training (fewer users) works better. Table~\ref{Table:Users} presents accuracy for deterministic elliptical quantization as a function of users. In contrast, our recognition algorithm is not sensitive to the number of users and has high accuracy for both user-dependent (fewer users) and user-independent (more users) training. 

\begin{figure}[!tb]
\centering
\includegraphics[height=1in, width=3.4in]{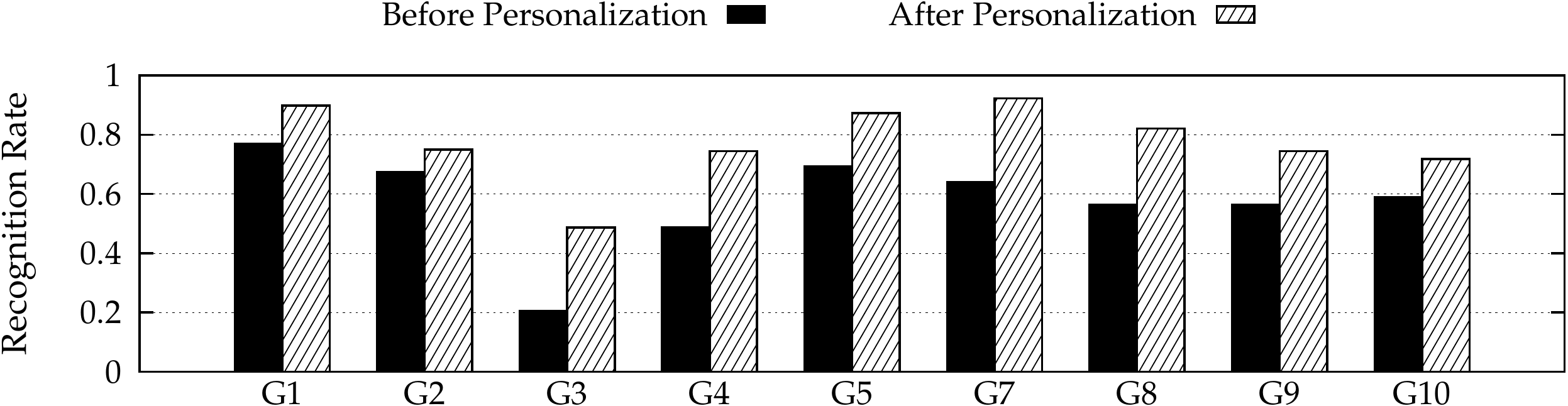}
\caption{\label{fig:Personalization} Classification accuracy of deterministic elliptical quantizer with personalization using \uncT.}
\end{figure}

\textbf{Frequency-based personalization} This section shows how our system easily incorporates additional sources of domain-specific information to improve accuracy.
Suppose the gesture recognition application trains with 20 gestures from 8 people. In deployment, the gesture recognition application detects that the user makes 10 gestures with equal probability, but very rarely makes the other 10 gestures. We prototype this scenario, by expressing the user-specific distribution of the 20 gestures as a probability distribution in the \uncT programming framework in the classification code. At classification time, the runtime combines this distribution over the gestures with the HMM to improve accuracy.  This figuration adds personalization to the deterministic elliptical quantization.  Figure~\ref{fig:Personalization} shows how the distribution of gestures preformed by a specific user improves gesture recognition accuracy from 10 to 20\% for each of the 10 gestures. Personalization could also be combined with statistical GMM.

\begin{table}[!tb]
\centering
\begin{tabular}{@{}p{3.5cm} p{3.5cm}@{}} 
\textsf{\textbf{Users}} &\textsf{\textbf{Accuracy}} 
\\ \midrule
2 & 82.71 \\
4 &  85.09\\
6 & 84.15\\
8 & 82.08 \\ \bottomrule \\ 
\end{tabular}
\caption{\label{Table:Users} Classification accuracy of deterministic elliptical quantizer  gestures 1, 3, 5, 7, 9 and 11 from the SW data set as a function of the number of users training the HMM.}
\end{table}

\begin{figure}[!tb]
\centering
\includegraphics[height=1in, width=3.5in]{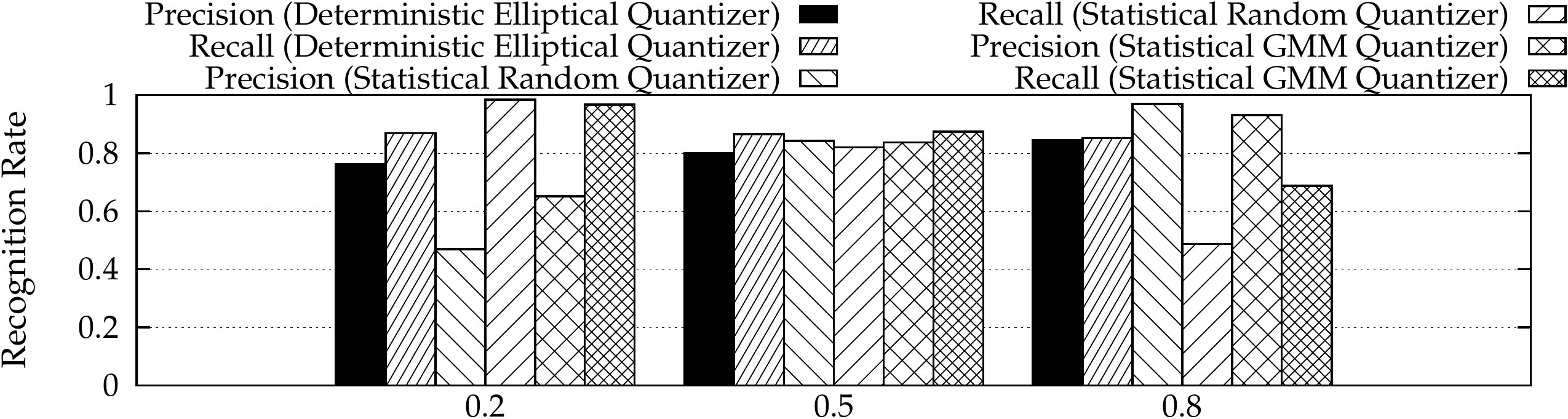}
\caption{\label{fig:Balance} Balancing false positives and false negatives with the statistical random and GMM quantizer on 10 SW gestures.}
~\\
\begin{scriptsize}
\begin{lstlisting}
public static bool Pr(this Uncertain<bool> source, double Prob = 0.5, double Alpha = 0.1);
 \end{lstlisting}
 \end{scriptsize}
 \caption{The default values for .Prob() inference calls.}
  \label{fig:ProbFunction}
~\\
\centering
\includegraphics[height=1in, width=3.5in]{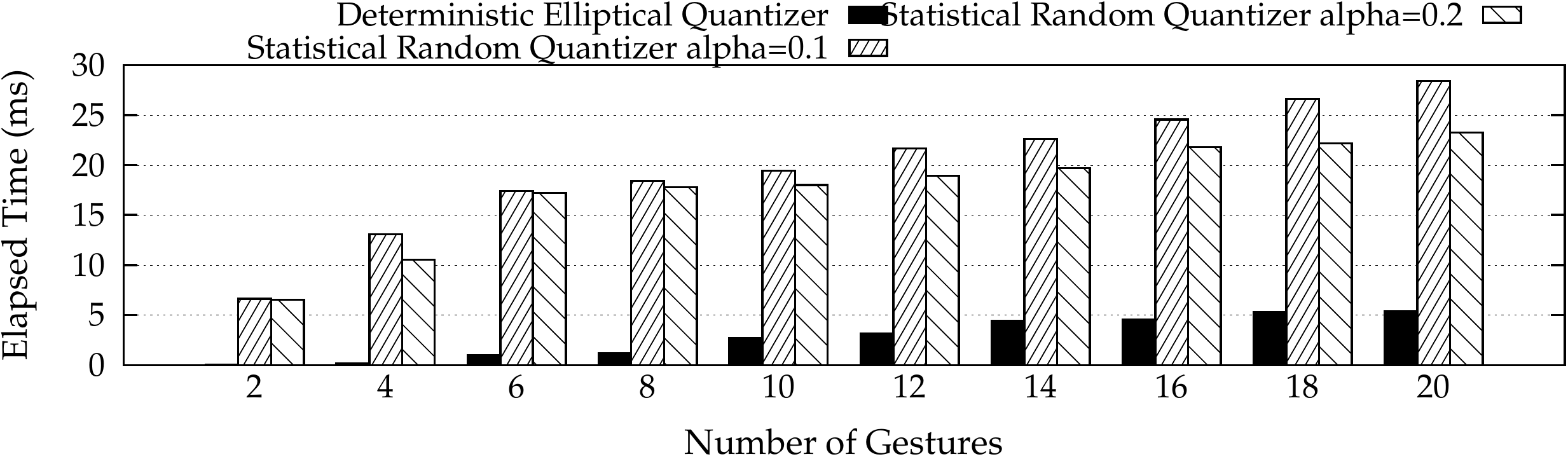}
\caption{\label{fig:CostOfRandomQuantizer} Time elapsed for classification in the statistical random and deterministic elliptical quantizer techniques.}
\end{figure}

\textbf{Balancing false positives and false negatives}
This experiment shows in more detail how the statistical quantization balances false positives with false negatives. In contrast, the deterministic elliptical quantizer always returns a classification with either a high probability (near 1) or low probability (near zero). Figure~\ref{fig:Balance} shows a case study of classification of 10 gestures from the SW data set. The figure shows the recognition rate of a gesture whose recognition rate for deterministic elliptical quantizer is 0.90 and for statistical random quantizer and is 0.87. However, for the statistical random quantizer the balance between false positives and false negatives occurs at a higher threshold (near 0.5), which means that changing the conditional threshold of the classifier can decrease false negatives. However in the deterministic elliptical quantizer, the balance between false positives and false negatives happens at a lower conditional threshold, which means that the probability of false negatives is always higher in this classifier.
\textbf{Cost of statistical random quantization}
While the statistical random quantizer gives us the flexibility of higher precision or recall, it incurs more recognition time. We show that this overhead is low in absolute terms, but high compared to a classifier that does not explore multiple options. Figure~\ref{fig:CostOfRandomQuantizer} graphs how much time it takes to recognize different number of gestures with the deterministic elliptical quantization and statistical random quantization techniques. On average the statistical random quantizer is 16 times slower at recognizing 2--20 different gestures, taking 23~ms to recognize 20 gestures and 6.5~ms for two gestures.
The statistical random quantizer uses \texttt{.Pr()} calls to invoke statistical hypothesis tests, and thus samples the computation many times.  Figure~\ref{fig:ProbFunction} shows the default value for the \texttt{.Pr()} function.
If we change the value of $\alpha$ for the statistical test from 0.1 to 0.2, the time overhead reduces from 28~ms to 23~ms. 
 If the system needs to be faster, statistical quantization trials are independent and could be performed in parallel. This additional overhead is very unlikely to degrade the user experience because in absolute terms, it is still much less than the 100\,ms delay that is perceptible to humans~\cite{perception:2016}.

\section{Conclusion} \label{Conclusion}
The promise of novel applications for sensing humans and machine learning
is only realizable
if, as a community, we help developers to use these tools correctly.
This paper
demonstrates that human gestures are very noisy and degrade the accuracy of machine learning models
for gesture recognition.  To help developers to more accurately
deal with gesture noise, we introduce probabilistic
quantization wherein gesture recognition finds the
most likely sequence of hidden states given a \emph{distribution} over
observations rather than a single observation. We express this new
approach using \uncT, a probabilistic programming system that
automates inference over probabilistic models. We demonstrate how
\uncT helps developers balance
false positives with false negatives at gesture recognition time,
instead of at gesture training time.  Our new gesture
recognition approach improves recall and precision over prior work on 25
gestures from 28 people.

\bibliographystyle{unsrt}{
\bibliography{bib/references}}

\end{document}